\documentclass{IEEEtran}

\usepackage{cite}
\usepackage{amsmath,amssymb,amsfonts}
\usepackage{algorithmic}
\usepackage{graphicx}
\usepackage{algorithm,algorithmic}
\usepackage[labelformat=simple]{subcaption}
\usepackage{hyperref}

\hypersetup{hidelinks=true}
\usepackage{textcomp}
\usepackage{dblfloatfix}
\def\BibTeX{{\rm B\kern-.05em{\sc i\kern-.025em b}\kern-.08em
    T\kern-.1667em\lower.7ex\hbox{E}\kern-.125emX}}
\begin{document}
\title{GPT Sonograpy: Hand Gesture Decoding from Forearm Ultrasound Images via VLM}
\author{Keshav Bimbraw, Ye Wang, \IEEEmembership{Senior Member, IEEE}, Jing Liu, Toshiaki Koike-Akino, \IEEEmembership{Senior Member, IEEE}
\thanks{K.B is with Worcester Polytechnic Institute, Worcester, MA 01609, USA 
(e-mail: kbimbraw@wpi.edu).}
\thanks{The authors are with Mitsubishi Electric Research Laboratories (MERL), 201 Broadway, Cambridge, MA 02139, USA 
(e-mail: \{bimbraw, yewang, jiliu, koike\}@merl.com).}
\thanks{This work was conducted while K.B was an intern at MERL.}
}

\maketitle

\begin{abstract}
Large vision-language models (LVLMs), such as the Generative Pre-trained Transformer 4-omni (GPT-4o), are emerging multi-modal foundation models which have great potential as powerful artificial-intelligence (AI) assistance tools for a myriad of applications, including healthcare, industrial, and academic sectors. Although such foundation models perform well in a wide range of general tasks, their capability without fine-tuning is often limited in specialized tasks. However, full fine-tuning of large foundation models is challenging due to enormous computation/memory/dataset requirements. We show that GPT-4o can decode hand gestures from forearm ultrasound data even with no fine-tuning, and improves with few-shot, in-context learning.
\end{abstract}


\section{Introduction} \label{sec:introduction}

\IEEEPARstart{L}{arge} language models (LLMs)\cite{zhao2023survey}, such as generative pre-trained transformers (GPTs)\cite{radford2018improving}, have recently emerged as powerful general assistance tools and exhibited tremendous capabilities in a wide range of applications.
LLMs are often configured with billions of parameters to capture linguistic patterns and semantic relationships in natural language processing, enabling text generation, summarization, translation, reasoning, question-answering, etc.

More recently, large multi-modal models (LMMs)\cite{wang2023large} with the capability to understand both natural language and other modalities, such as images and sounds, have offered new opportunities for biomedical applications.
For example, it was demonstrated that large vision-language models (LVLMs) such as GPT-4o\cite{shahriar2024putting} and LLaVa\cite{liu2024improved} could be a viable tool for medical applications\cite{zhang2024latest}, including surgical oncology\cite{zhu2024openai} and radiology diagnosis\cite{sonoda2024diagnostic, oura2024diagnostic, cesur2024empowering}.
We examine the capabilities of GPT-4o for sonography\cite{kremkau2015sonography}, to analyze and decode ultrasound images.

\begin{figure}[t!]
    \centering
    \includegraphics[width=250pt]{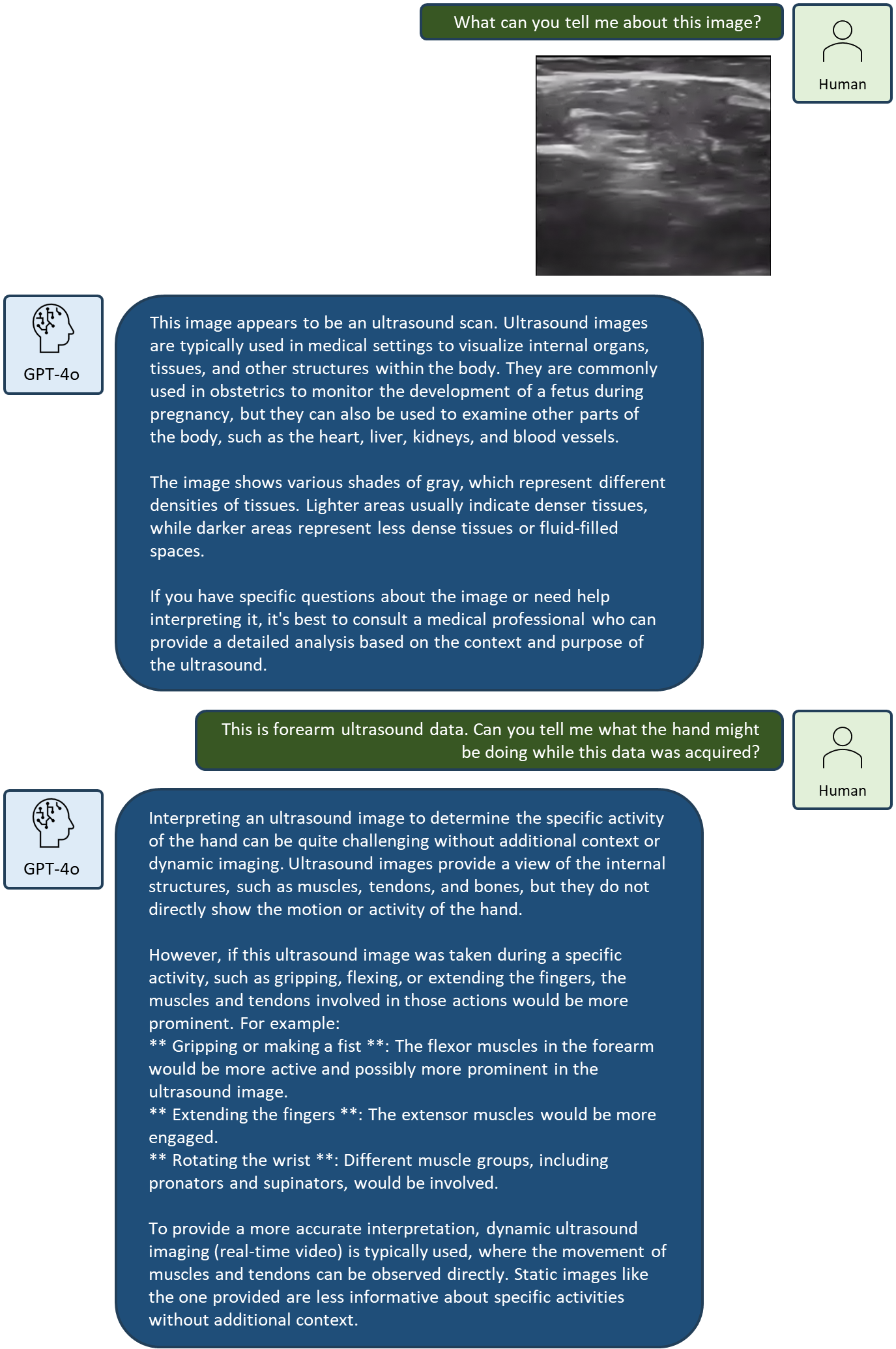}
    \caption{Conversation with GPT-4o that motivated us to use the VLM for ultrasound image decoding.}
    \label{fig:motivation}
\end{figure}

\begin{figure*}[hbt!]
    \centering
    \includegraphics[width=350pt]{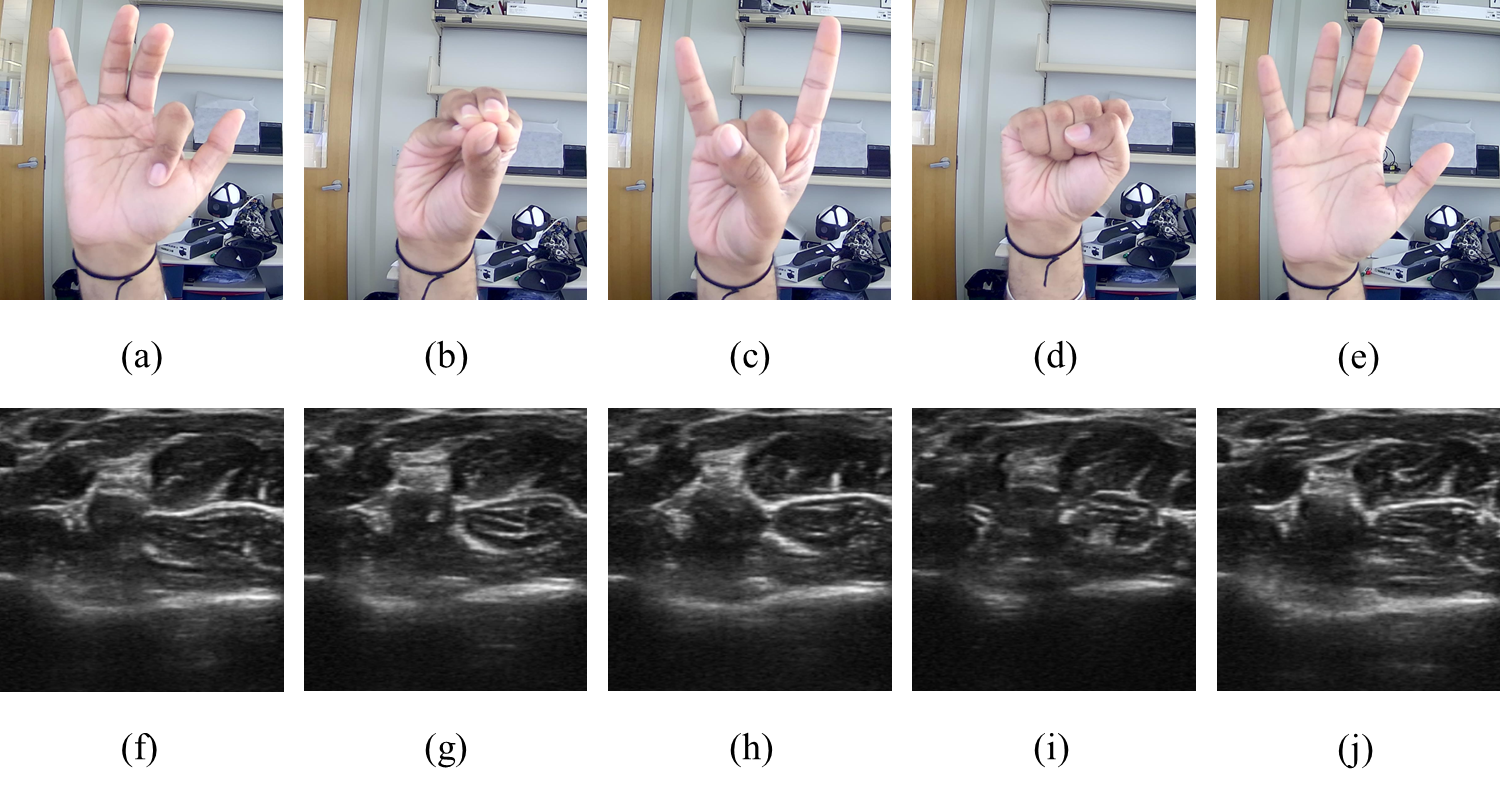}
    \caption{Hand gestures (a through e) and the corresponding forearm ultrasound image (f through j) from subject 1. 
    (a) and (f): Index flexion; (b) and (g): all pinch; (c) and (h) hand horns; (d) and (i) fist; (e) and (j): open hand.}
    \label{fig:gestures}
\end{figure*}

Musculoskeletal ultrasound is a non-invasive and non-radiative imaging technique that uses ultrasound waves to visualize muscles, tendons, ligaments, and joints. For instance, ultrasound measurements can be used to visualize the anatomical aspects of the forearm, to estimate hand gestures\cite{bimbraw2023simultaneous, mcintosh2017echoflex}.
This is applicable to several domains, such as control of prosthetic hands\cite{yin2022wearable}, teleoperation of robotic grippers\cite{bimbraw2020towards}, and controlling virtual reality interfaces \cite{bimbraw2023leveraging}.
In particular, modern deep learning methods have shown improved performance to estimate different hand gestures\cite{bimbraw2024mirror}.
It is highly expected that the use of LVLMs like GPT-4o to classify ultrasound images can provide a lot more information through human readable explanations of the model's predictions, which aids understanding of the reasoning behind gesture recognition.
In addition, contextual information can be potentially leveraged to improve the classification performance.  

Although the pre-trained LVLMs work well for a general task, its performance is often limited for specialized tasks such as biomedical dataset.
Given such a dataset, fine-tuning can greatly improve the performance for downstream tasks in general.
Nevertheless, fine-tuning LVLMs is challenging due to the substantial amount of labelled data required\cite{zhai2024fine}. Additionally, it demands significant computational resources and time.
Therefore, it is more practical and cost-effective to consider using the pre-trained LVLMs without fine-tuning but with prompt tuning\cite{lester2021power} or in-context learning (ICL)\cite{brown2020language}.
ICL does not modify the pre-trained LVLMs, but instead adds some task-specific examples to the input context to improve the performance of generating the desired responses.

In this work, we show that we can leverage GPT-4o to classify ultrasound images using a few-shot ICL strategy. 
We demonstrate that providing some labelled examples to the LVLM significantly improves its performance for forearm ultrasound-based gesture recognition. 
This opens up exciting applications for LVLMs in medical imaging. 
The contributions of this paper are summarized as follows.
\begin{itemize}
\item We examine the capability of LVLMs for sonography diagnosis.
\item We use GPT-4o to analyze forearm ultrasound images for hand gesture decoding.
\item We demonstrate that GPT-4o can achieve high accuracy of over 70\% for cross-subsession experiments to classify hand gesture even without any fine-tuning.
\item We show that the few-shot ICL strategy is substantially effective to improve the classification accuracy.
\item We provide some discussions on cross-subject transfer, prompt engineering, and image augmentations.
\end{itemize}

\section{Motivation}

LVLMs have the capability to handle tasks that involve both images and texts. 
They have proven to be useful for understanding medical image data, especially with extensive fine-tuning \cite{sonoda2024diagnostic}. 
Since full fine-tuning of LVLMs requires substantial computational resources, we first examined to see how GPT-4o would perform without fine-tuning. 
GPT-4o was provided a forearm ultrasound image, and asked a simple question ``What can you tell me about this image?''. 
The LVLM was able to identify that it is an ultrasound image, and gave some additional information about generic ultrasound images and their visual properties. 
We then examined whether it could infer some additional information when it is given some context. 
To this end, a follow-up question was asked: ``This is forearm ultrasound data. 
Can you tell me what the hand might be doing while this data was acquired?''. 
The LVLM gave some more information about physiology of hand movement and how different hand movements would lead to different ultrasound images. 
The full conversation can be seen in Fig.~\ref{fig:motivation}. 

This motivated us to experiment with GPT-4o to see if it could classify forearm ultrasound images corresponding to different hand movements. 
We are also interested in evaluating its performance while varying the amount of data and context that it is exposed to.

\begin{figure}[t!]
    \centering
    \includegraphics[width=250pt]{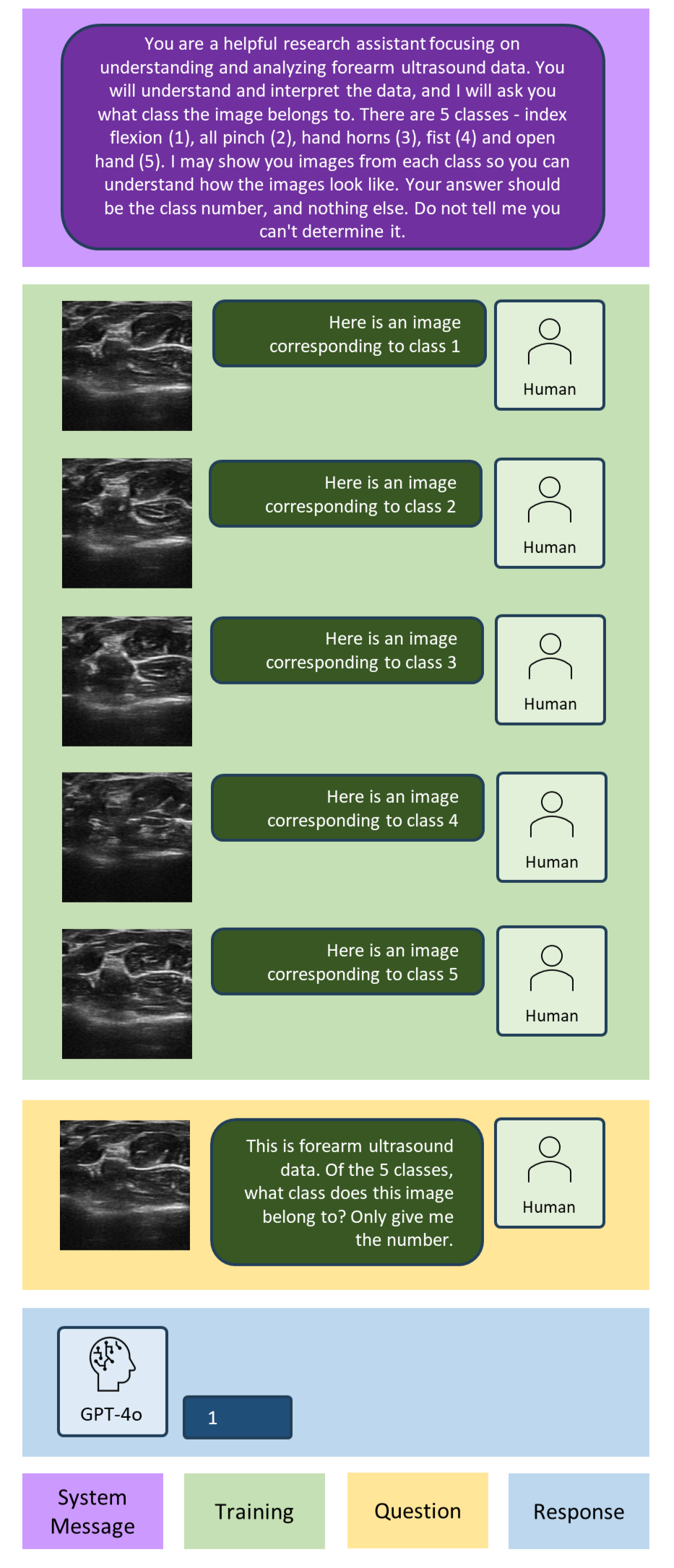}
    \caption{Conversation with GPT-4o for forearm ultrasound classification based on 1-shot learning.}
    \label{fig:conversation}
\end{figure}

\section{Methodology}

For this study, ultrasound data was acquired from 3 subjects. 
The study was approved by the institutional research ethics committee (IRB reference number 23001). 
Written informed consent was given by the subjects before data acquisition. 
Per subject, data was acquired for 5 hand gestures as shown in Figs.~\ref{fig:gestures}: 
(1) index flexion; (2) all pinch; (3) hand horns; (4) fist; and (5) open hand. 
These are based on activities of daily living and the chosen gestures are a subset of the dataset in \cite{bimbraw2023simultaneous}.

\begin{figure*}[hbt!]
    \centering
    \includegraphics[width=500pt]{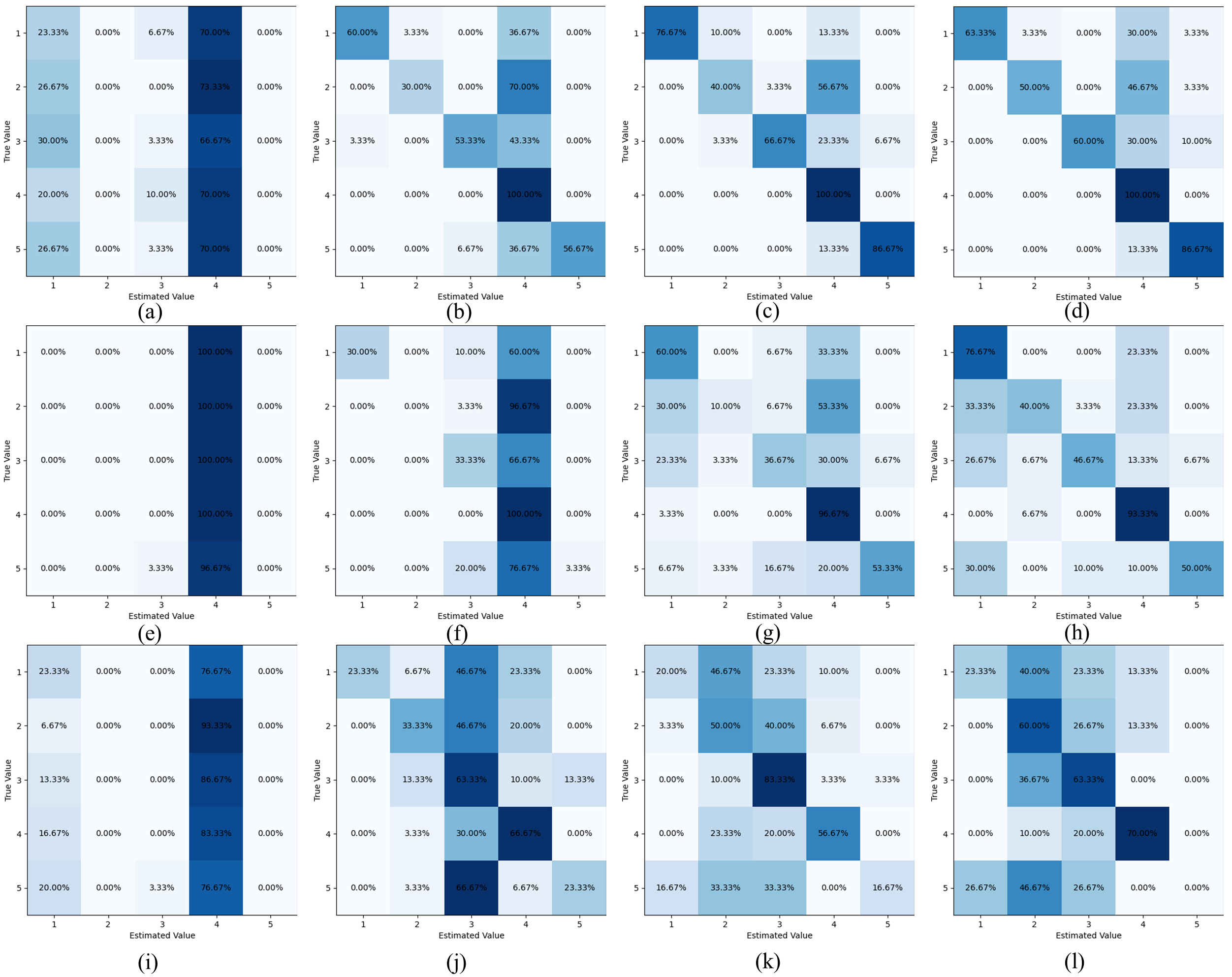}
    \caption{Confusion matrices for within-session (a--d), cross-session (e--h), and randomized cross-session (i--l) experiments summed over the three subjects for: 
    0-shot (a, e, and i), 1-shot (b, f, and j), 2-shot (c, g, and k), and 3-shot (d, h, and l) strategies.}
    \label{fig:confusion}
\end{figure*}

\subsection{Data Acquisition}

The ultrasound data was acquired using a Sonostar 4L linear palm Doppler ultrasound probe\cite{sonostar}. 
A custom-designed 3D-printed wearable was strapped onto the subject's forearm. 
The data from the probe was streamed to a Windows system over Wi-Fi, and screenshots of the ultrasound images were captured using a custom Python script. 
The 4L linear probe has 80 channels of ultrasound data, and the post-processed beamformed B-mode data is obtained, from which $350 \times 350$-pixel images are acquired.

For each subject, 5 sessions of data were collected. 
In each session, subjects performed a sequence of 5 gestures. 
Within each session, this sequence was repeated 4 times, resulting in 20 sub-sessions. 
For our study, we analyzed 10 frames per sub-session, resulting in a total of 1000 images (i.e., 20 sub-sessions, 10 frames/sub-session, 5 gestures) per subject.

\subsection{Large Vision-Language Model (LVLM)}

We use GPT-4o\cite{shahriar2024putting} as one of state-of-the-art LVLMs. 
GPT-4o is a multi-modal generative pre-trained transformer designed by OpenAI. 
It is said that GPT-4o uses more than 175 billion parameters. 
GPT-4o integrates texts and images in a single model, enabling it to handle multiple data types simultaneously. 
This multi-modal approach enhances accuracy and responsiveness in human-computer interactions. 
For inference, Azure OpenAI module within OpenAI's Python library was used \cite{githubGitHubOpenaiopenaipython}. 
Azure cloud computing was used within a Linux system with Python 3.11 for scripting.

The image data needs to be converted to a text format so that GPT-4o can understand it. 
For this study, the ultrasound image data was encoded to base64 using the Python base64 library, resulting in a text-based representation suitable for transmission or embedding \cite{key}. 
The ultrasound image is represented as a long string of text upon encoding.

\subsection{GPT-4o Prompts}

The conversation flow we use is described in Fig.~\ref{fig:conversation}. 
To effectively utilize GPT-4o, we designed the conversation as follows.

\subsubsection{System Message} 
We began with a system message to set context and guidelines for the conversation. 
GPT-4o was informed that it would serve as a helpful research assistant and will assist in classifying hand gestures using forearm ultrasound data.

\subsubsection{In-Context Learning (ICL)} 
We used an ICL strategy which  provides training examples in contexts. 
We use a few forearm ultrasound image samples along with the class labels for the in-context examples to assist GPT-4o for specialized classification tasks. 
Note that ICL does not involve any `learning' procedure such as fine-tuning, adaptation, or post-training.

\subsubsection{Query for Classification} 
The GPT-4o was then asked to predict the hand gesture class based on the given ultrasound image. 
It was explicitly instructed to provide just the class number, which can be saved for further analysis.

\section{Experimental Setup}

The performance was evaluated with few-shot in-context strategies: 0-shot; 1-shot; 2-shot; and 3-shot ICL. 
Two experiments were carried out: within-session analysis and cross-session analysis. 
For the former, for a given subject, of the 40 images per class in session 1, the last sub-session (last 10 images) were used for evaluation, and the remaining were used for training. 
For the latter, the last sub-session of session 5 was used for evaluation, while the remaining data was used as ICL training samples. 
For the three different experiments, different data was used for training and evaluation. 
For 0-shot strategy, the LVLM was shown images in the test set directly and asked what class out of the 5 it belonged to. 

\subsection{Within-Session Analysis}

For the 1, 2, and 3-shot strategies, the data-split is described below.
\subsubsection{1-Shot}
The first image per class from sub-session 1 was shown to the model along with the class label before asking the question. 
This leads to a total of 5 images and their corresponding class-labels shown. 
This can be seen in Fig.~\ref{fig:conversation}.
\subsubsection{2-Shot}
The first two images per class from sub-session 1 were shown to the model along with the class labels, leading to a total of 10 images shown.
\subsubsection{3-Shot}
The first image per class from sub-sessions 1, 2, and 3 were shown to the model along with the class label, leading to a total of 15 images shown.

\subsection{Cross-Session Analysis}

For the 1, 2, and 3-shot strategies, the data-split is described below.
\subsubsection{1-Shot}
The first image per class from sub-session 1 was shown to the model along with the class label before asking the question. 
This leads to a total of 5 images and their corresponding class-labels shown. 
The training data shown in similar to the within-session experiment.
\subsubsection{2-Shot}
The first image per class (sub-session 1) from sessions 1 and 2 were shown to the model along with the class labels, leading to a total of 10 images shown.
\subsubsection{3-Shot}
The first image from sub-session 1 per class from sessions 1, 2, and 3 were shown to the model along with the class label, leading to a total of 15 images shown.

\subsection{Evaluation Metrics}

To evaluate the performance, the predicted class labels from GPT-4o were compared to the true values. 
Classification accuracy was used as a metric for evaluating the performance. 
Confusion matrices were used to visualize the performance for different scenarios. 
Precision, recall and F1 scores were also calculated for each confusion matrix.

\section{Results}

This section provides the results for within-session and cross-session experiments for 0-shot, 1-shot, 2-shot, and 3-shot ICL strategies. 

\subsection{Within-Session Experiment}

The confusion matrix, summed over the three subjects for the within-session experiment can be seen in Fig.~\ref{fig:confusion}(a)--(d) for 0, 1, 2, and 3-shot strategies respectively. 

\begin{figure*}[h!tbp]
    \centering
    \begin{subfigure}{0.32\linewidth}    \includegraphics[trim=80 0 80 0, clip, width=\linewidth]{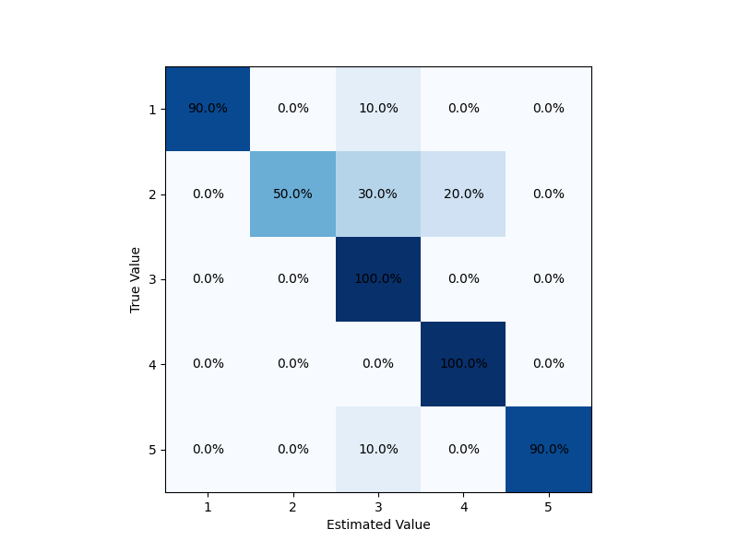}
    \caption{Baseline (Accuracy: 86\%)}
    \label{fig:d1}
    \end{subfigure}
    \begin{subfigure}{0.32\linewidth}    \includegraphics[trim=80 0 80 0, clip, width=\linewidth]{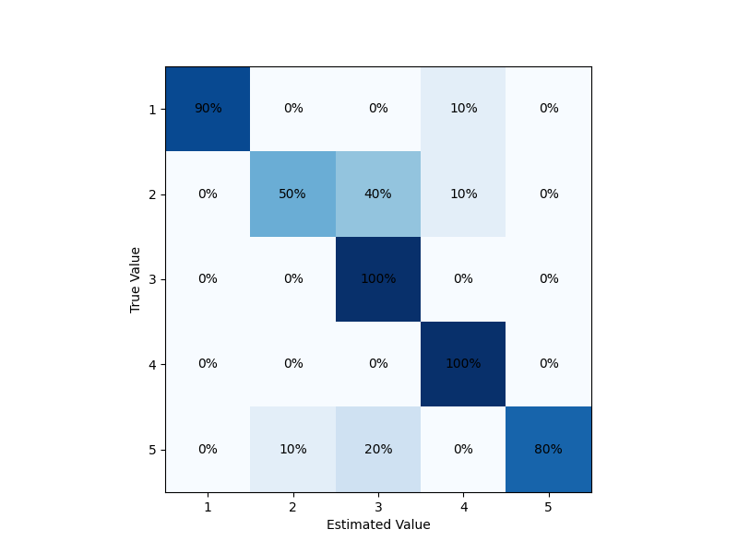}
    \caption{Low-descriptive (Accuracy: 82\%)}
    \label{fig:d2}
    \end{subfigure}
    \begin{subfigure}{0.32\linewidth}    \includegraphics[trim=80 0 80 0, clip, width=\linewidth]{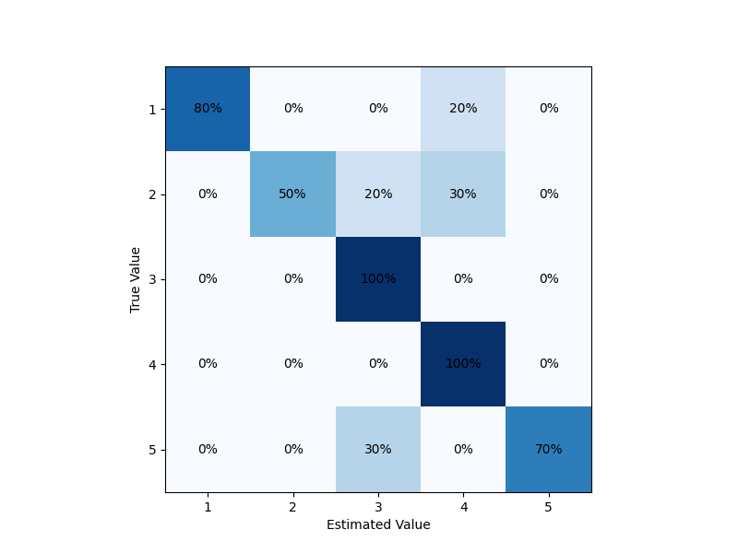}
    \caption{High-descriptive (Accuracy: 80\%)}
    \label{fig:d3}
    \end{subfigure}

\caption{Confusion matrices with different prompts (within-session, subject 1, 1-shot).}
    \label{fig:d0}
\end{figure*}

The classification accuracy, along with the precision, recall, and F1 scores are summarized in table~\ref{table1}. 

\begin{table}[t]
\caption{Within-session experiment results}
\centering
\begin{tabular}{lllll}
\hline
 & Accuracy & Precision & Recall & F1 Score \\ \hline
0-shot & 0.193 & --- & 0.193 & --- \\
1-shot & 0.600 & 0.817 & 0.600 & 0.618 \\ 
2-shot & \textbf{0.740} & 0.826 & \textbf{0.753} & \textbf{0.756} \\ 
3-shot & 0.720 & \textbf{0.846} & 0.720 & 0.731 \\ \hline
\end{tabular}
\label{table1}
\end{table}

\begin{table}[t]
\caption{Cross-session experiment results}
\centering
\begin{tabular}{lllll}
\hline
 & Accuracy & Precision & Recall & F1 Score \\ \hline
0-shot & 0.200 & --- & 0.200 & --- \\ 
1-shot & 0.333 & --- & 0.333 & --- \\ 
2-shot & 0.513 & 0.588 & 0.513 & 0.479 \\ 
3-shot & \textbf{0.613} & \textbf{0.688} & \textbf{0.613} & \textbf{0.605} \\ \hline
\end{tabular}
\label{table2}
\end{table}

Table~\ref{table3} shows the classification accuracy averaged over three subjects for within-session experiment. 
For 0-shot strategy, the average classification accuracy was 19.3\% ($\pm$1.0\%). 
For 1-shot, 2-shot and 3-shot strategies, we achieved 60.0\% ($\pm$15.9\%), 74.0\% ($\pm$12.0\%), and 72.0\% ($\pm$16.0\%) respectively. 
It clearly demonstrates that in-context examples can significantly improve the classification accuracy even without fine-tuning the pre-trained LVLM. 
A slight decline of 2 percentage points is observed when the training examples increase from 2 to 3 per class.
It may be within a statistical fluctuation due to the small number of test samples. 

\subsection{Cross-Session Experiment}

The confusion matrix, summed over the three subjects for the cross-session experiment can be seen in Figs.~\ref{fig:confusion}(e)--(h) for 0, 1, 2, and 3-shot strategies respectively. The classification accuracy, along with the precision, recall, and F1 scores are summarized in table~\ref{table2}. 

\begin{table}[t]
\caption{Average accuracy comparisons for within-session, cross-session, and randomized cross-session experiments}
\centering
\begin{tabular}{cccc}
\hline
 & Within-Session & Cross-Session 
 & Randomized \\ 
 \hline
0-Shot & 0.193 ($\pm 0.012$) & 0.200 ($\pm 0.000$) & 0.213 ($\pm 0.031$) \\ 
1-Shot & 0.600 ($\pm 0.159$) & 0.333  ($\pm 0.167$) & 0.420 ($\pm 0.106$) \\ 
2-Shot & \textbf{0.740} ($\pm 0.120$) & 0.513 ($\pm 0.155$) & \textbf{0.453} ($\pm 0.050$) \\ 
3-Shot & 0.720 ($\pm 0.160$) & \textbf{0.613} ($\pm 0.223$) & 0.433 ($\pm 0.042$) \\ \hline
\end{tabular}
\label{table3}
\end{table}

For 0-shot case in Fig.~\ref{fig:confusion}(a), the classification accuracy is comparable to a random guess because of 5 classes. 
For 1-shot strategy in Fig.~\ref{fig:confusion}(b), it was 52\%. 
For 2-shot in Fig.~\ref{fig:confusion}(c), it was 56\%, which increased to 70\% for 3-shot case as in Fig.~\ref{fig:confusion}(d). 
This trend is encouraging since increasing the number of in-context samples can improve the performance of GPT-4o to classify forearm ultrasound images to predict the hand gestures they correspond.  

This was repeated for subjects 2 and 3. 
Table~\ref{table3} shows the classification results averaged over the three subjects. 
For 0-shot strategy, the average classification accuracy was 20.0\% ($\pm$0.0\%). 
For 1-shot, 2-shot and 3-shot strategies, it was obtained to be 33.3\% ($\pm$16.7\%), 51.3\% ($\pm$15.5\%), and 61.3\% ($\pm$22.3\%) respectively. 
These results show a clear improvement in the classifier performance for an increasing number of in-context samples. 
It was interesting to observe that the standard deviation increases sharply as the number of training examples increases from 2 to 3 per class.

\begin{figure}[t]
    \centering
    \includegraphics[width=250pt]{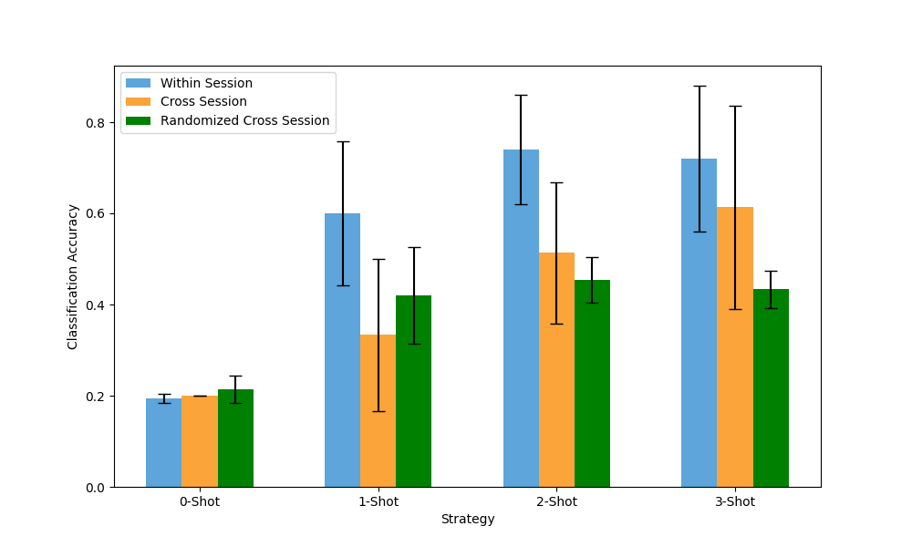}
    \caption{Average classification accuracy for within-session and cross-session experiments, visualizing results in Table~\ref{table3}.}
    \label{fig:table3}
\end{figure}

The results for the case where the input samples were picked randomly from the training data is shown in Table~\ref{table3}. 
While the performance with in-context learning was better than 0-shot case, it was worse than non-randomized case. 
Increasing the number of training samples did not clearly improve the average classification across subjects. 
For 0-shot strategy, the classification accuracy was 21.3\% ($\pm$3.0\%). 
For 1-shot strategy, the average classification accuracy was 42.0\% ($\pm$10.6\%). 
For 2-shot strategy, the average classification accuracy was 45.3\% ($\pm$5.0\%). 
And for 3-shot strategy, the average classification accuracy was 43.3\% ($\pm$4.2\%). 
The results are summarized in Fig.~\ref{fig:table3}.

\begin{figure}[h!]
    \centering
    \includegraphics[width=225pt]{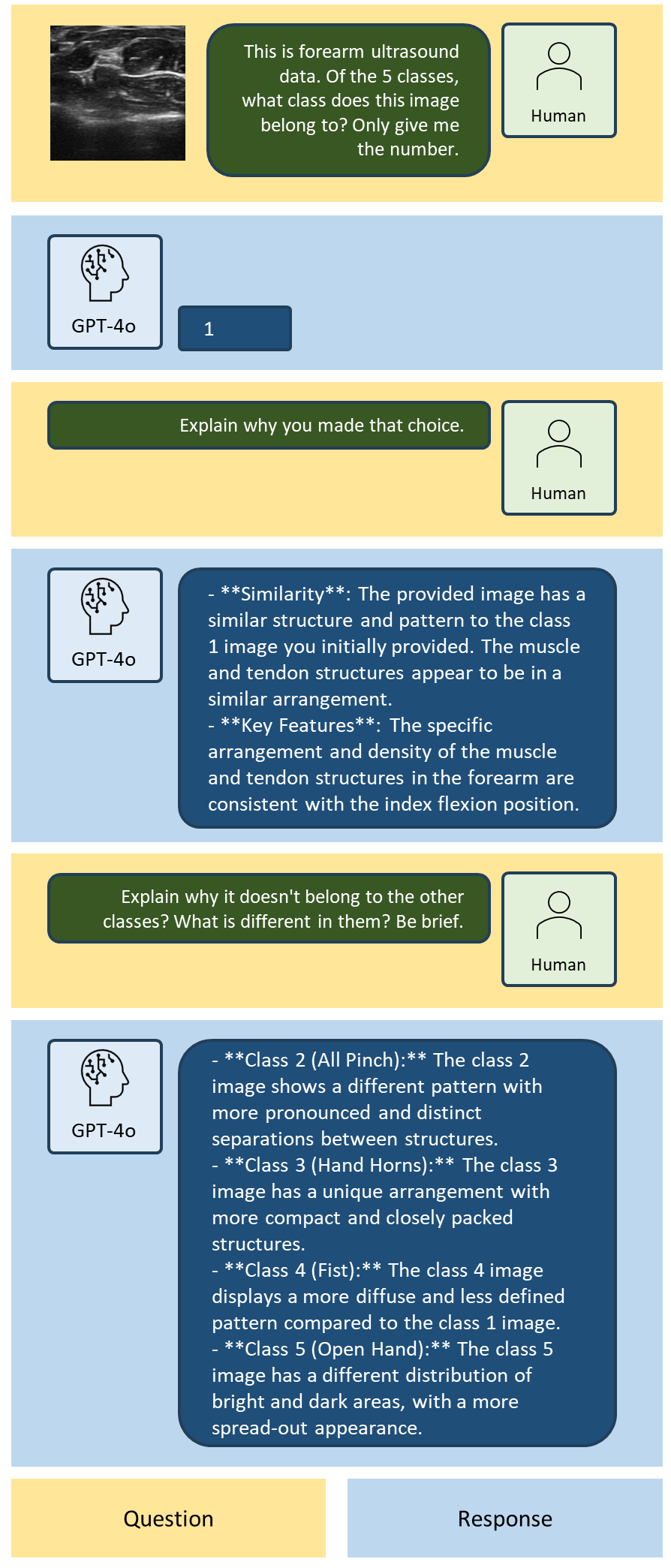}
    \caption{Conversation with GPT-4o as a follow up to the 1-shot conversation in Fig.~\ref{fig:conversation} to demonstrate its reasoning capabilities.}
    \label{fig:reasoning}
\end{figure}

\section{Discussion}
Several additional experiments were carried out for within-session data from subject 1 to understand GPT-4o's performance and reasoning. All these experiments were done for a 1-shot strategy. The baseline confusion matrix is shown in Fig.~\ref{fig:d1}. For this case, the accuracy is 86\%, with the macro average precision, recall, and F1 scores being 0.9, 0.86, and 0.85, respectively. 

\subsection{Results with different prompts}
We wanted to see how GPT-4o would perform with prompts less and more descriptive than the prompts shown in Fig.~\ref{fig:conversation}. 
\subsubsection{Less descriptive information}
For this experiment, we did not provide the system message. 
And for training, we only stated the class label with the image. 
As the question, we just asked `What class does the image belong to? Only give the class number.' 
With this minimal information, the confusion matrix obtained is shown in Fig.~\ref{fig:d2}. 
For this case, the accuracy is 82\%, with the macro average precision, recall, and F1 scores being 0.86, 0.82, and 0.82, respectively. 
It was interesting to see that there was only a decline of 4\% in the classification accuracy from the baseline of Fig.~\ref{fig:d1}, meaning that we can provide it a lot less information without compromising significantly on the accuracy.

\subsubsection{More descriptive information}

For this experiment, we provided a lot more contextual information to GPT-4o both in the system message, as well as in the final question. 
We mentioned that it should focus on the arrangement of regions with different brightness. 
We also mentioned that the anatomical and physiological properties visualized in the ultrasound image are distinct for different hand gestures. 
The confusion matrix is shown in Fig.~\ref{fig:d3}. 
For this case, the accuracy is 80\%, with the macro average precision, recall, and F1 scores being 0.87, 0.8, and 0.8, respectively. 

It was interesting to see that providing so much extra information did not really help improve the performance. 
Rather, it decreased the performance compared to the less descriptive information case by 2\%.

\subsection{Reasoning ability}

With the flow shown in Fig.~\ref{fig:conversation}, we wanted to understand why GPT-4o made that particular estimation. Fig.~\ref{fig:reasoning} shows the user asking questions to GPT-4o, and it answering why it made that particular estimation compared to the other classes. 
Based on this conversation, we can make the following conclusions. 

\subsubsection{Logical Coherence} 
GPT-4o demonstrates a structured approach to reasoning, with each successive step logically following the previous one. 
This indicates an ability to maintain logical consistency.

\subsubsection{Contextual Understanding} 
The model incorporates context into its reasoning, ensuring that decisions are relevant to the given scenario. 
It takes into consideration the information provided during training, as well as in the system message.

\subsubsection{Decision-Making} 
GPT-4o was able to express why the image does not belong to the other classes. 
It provides a clear delineation between the the different classes, such as for class 5 (open hand), it stated that there is a different distribution of bright and dark areas with more spread out experience, and hence, the image does not belong to class 5. 

While the model's reasoning is not fully trustworthy and VLMs are prone to hallucinations \cite{liu2024survey}, it is encouraging to see that VLMs like GPT-4o can be used to understand better why it made a particular prediction. 
More effective conversations with contextual clues may improve its performance.

\begin{figure}[t]
    \centering
    \includegraphics[width=150pt]{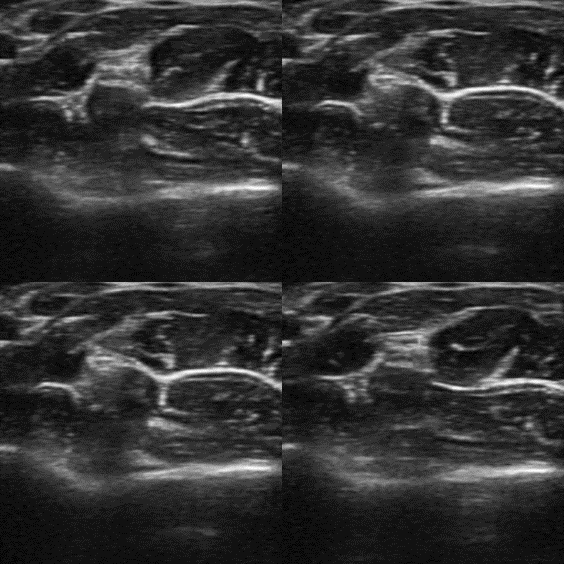}
    \caption{Stacked ultrasound images for class 1 with ultrasound image frames taken at different times.}
    \label{fig:combined}
\end{figure}

\begin{figure}
    \centering
    \begin{subfigure}{\linewidth}
    \includegraphics[width=250pt]{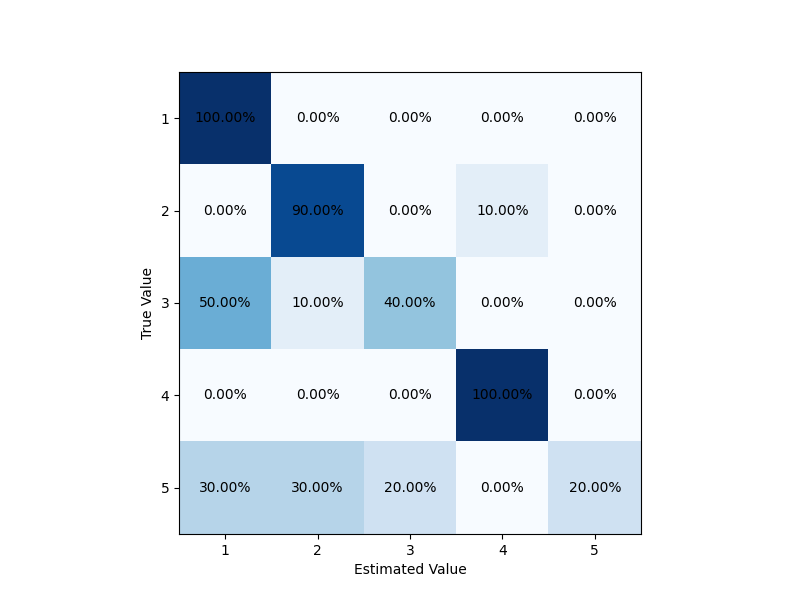}
    \caption{Stacked 2-frame (Accuracy: 78\%)}
    \label{fig:d_s_2}
    \end{subfigure}
    
    \begin{subfigure}{\linewidth}
    \includegraphics[width=250pt]{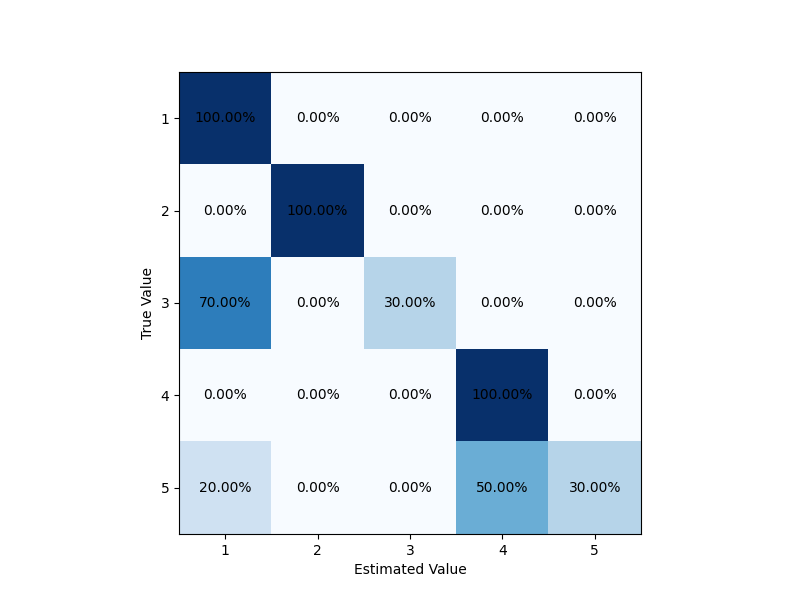}
    \caption{Stacked 4-frame (Accuracy: 72\%)}
    \label{fig:d_s_4}
    \end{subfigure}
    \caption{Confusion matrix given stacked ultrasound images.}
\end{figure}

\subsection{Different input formats}

Radiologists often look at stacked medical images to understand medical image data. 
This is done especially with time-varying data to visualize how the physiological features change with time \cite{radiopaediaStacksRadiology}. 
We wanted to see how GPT-4o would perform for different stacks of ultrasound images. Fig.~\ref{fig:combined} shows a stacked image sample with 4 ultrasound image frames.

\subsubsection{Two images as input}

Using two stacked ultrasound frames as input for 1-shot strategy, instead of one image per class, 1 image with two ultrasound frames corresponding to the class were shown. 
This can be visualized in the top row of Fig.~\ref{fig:combined}. 
The classification results are shown in Fig.~\ref{fig:d_s_2}. 
For this case, the accuracy is 78\%, with the macro average precision, recall, and F1 scores being 0.83, 0.78, and 0.77, respectively.

\subsubsection{Four images as input}

Using 4 stacked ultrasound frames as input for 1-shot strategy, instead of one image per class, 1 image with 4 ultrasound frames corresponding to the class were shown. 
This can be visualized in Fig.~\ref{fig:combined}. 
The classification results are shown in Fig.~\ref{fig:d_s_4}. 
For this case, the accuracy is 72\%, with the macro average precision, recall, and F1 scores being 0.84, 0.72, and 0.68, respectively. 

Although more training samples are provided by stacking frames, the classification accuracy was degraded. It may be because the image format is different for the testing image and the relative image resolution is lower when stacked. We believe that the performance can be improved by better designing prompts.

\subsection{Future work}

We conducted experiments to understand capabilities of GPT-4o for hand gesture classification based on forearm ultrasound data. 
We explored some interesting features of using VLMs for this task. 
Future work would include extensive cross validation analysis, in addition to acquiring data from more subjects. 
More rigorous prompt engineering should be considered as well.
We are also interested in exploring VLM's cross-subject generalizability for medical image datasets.  
In addition, the comparison to retrieval augmented generation (RAG)\cite{gao2023retrieval} and parameter efficient fine-tuning (PEFT)\cite{han2024parameter} methods should follow.

\section{Conclusions}

In this work, we show that we can use a large vision-language model (LVLMs), GPT-4o as a powerful AI assistance tool for understanding and interpreting forearm ultrasound data. 
We show that by providing some examples of ultrasound images, we can improve its performance for hand gesture classification based on forearm ultrasound data. 
For within-session performance, we show that the average gesture classification accuracy reached 74.0\% for 5 hand gestures with just 2 training samples, and for cross-session performance, it reached 61.3\% for just 3 training samples per class. 
Our approach can be used in cases where full-fine tuning of these models is challenging because of enormous compute/memory/dataset requirements. 
This research opens up exciting avenues for research in utilizing large vision-language models for medical imaging.

\section*{References}

\bibliography{refs}

\end{document}